\documentclass[times, 10pt]{article}
\usepackage{amsmath,amsthm,amsfonts,amssymb,amstext,latexsym,algorithm,algorithmic,graphicx}
\usepackage{times,subfigure}
\usepackage[sort&compress]{natbib}

\bibpunct{(}{)}{;}{a}{,}{,}

\newcommand{\R}{\mathbb{R}}				    
\newcommand{\BP}{\begin{proof}}	            
\newcommand{\EP}{\end{proof}}		        

\makeatletter
\floatstyle{\ALG@floatstyle}
\newfloat{protocol}{htbp}{loa}
\floatname{protocol}{Protocol}
\makeatother

\newtheorem{theorem}{Theorem}
\newtheorem*{theorem*}{Theorem}

\newtheorem{lemma}{Lemma}
\newtheorem{corollary}{Corollary}

\theoremstyle{definition}

\newtheorem{remark}{Remark}

\begin{document}

\title{Competing with Gaussian linear experts}
\date{}

\author{Fedor Zhdanov and Vladimir Vovk\\
Computer Learning Research Centre,\\
Department of Computer Science,\\
Royal Holloway, University of London,\\
Egham, Surrey, TW20 0EX, UK\\
\{fedor,vovk\}@cs.rhul.ac.uk\\
}

\maketitle
\thispagestyle{empty}

\begin{abstract}
We study the problem of online regression. We do not make any assumptions about input vectors or outcomes. We prove a theoretical bound on the square loss of Ridge Regression. We also show that Bayesian Ridge Regression can be thought of as an online algorithm competing with all the Gaussian linear experts. We then consider the case of infinite-dimensional Hilbert spaces and prove relative loss bounds for the popular non-parametric kernelized Bayesian Ridge Regression and kernelized Ridge Regression. Our main theoretical guarantees have the form of equalities.
\end{abstract}

\section{Introduction}\label{sec:introduction}
In the online prediction framework we are provided with some input at each step and try to predict an outcome using this input and information from previous steps \citep{CesaBianchi2006}. In a simple case in statistics, it is assumed that each outcome is the value, corrupted by Gaussian noise, of a linear function of input.

In competitive prediction the learner compares his loss at each step with the loss of any expert from a certain class of experts instead of making statistical assumptions about the data generating process. Experts may follow certain strategies. The learner wishes to predict almost as well as the best expert for \emph{all} sequences.

Our main result is Theorem~\ref{thm:main} in the next section,
which compares the cumulative weighted square loss of Ridge Regression
applied in the on-line mode
with the regularized cumulative loss of the best linear predictor.
The power of this result can be best appreciated by looking at the range of its implications,
both known and new.
For example, Corollary~\ref{cor:upperbound} answers the question
asked by several researchers, see \citet{VovkCOS},
whether Ridge Regression has a relative loss bound
with the regret term of the order $\ln T$ under the square loss function,
where $T$ is the number of steps and the outcomes are assumed bounded;
this corollary (as well as all other implications stated in Section~\ref{sec:protocol})
is an explicit inequality rather than an asymptotic result.
Theorem~\ref{thm:main} itself is much stronger,
stating an equality
rather than inequality
and not assuming that the outcomes are bounded.
Since it is an equality, it unites upper and lower bounds on the loss.
It appears that all natural bounds on the square loss of Ridge Regression can be
easily deduced from our theorem; we give some examples in the next section.

Most of previous research in online prediction considers experts that disregard
the presence of noise in observations. We consider experts predicting a
distribution on the outcomes. We use
Bayesian Ridge Regression and prove that it can predict as well as the best
regularized expert; this is our Theorem~\ref{thm:RRlogloss}. The loss in this theoretical guarantee is the logarithmic
loss. The algorithm that we apply was first used
by \citet{DeSantis1988}
and similar bounds to ours were obtained by \citet{Kakade2004,Kakade2005}.
Theorem~\ref{thm:RRlogloss} is later used to deduce Theorem~\ref{thm:main}.
Ridge Regression predicts the mean of the Bayesian Ridge Regression predictive distribution,
and the logarithmic loss of Bayesian Ridge Regression is close to scaled square loss
of Ridge Regression.

We extend our main result to the case of infinite dimensional
Hilbert spaces of functions. The algorithm used becomes an analogue of
non-parametric Bayesian methods. From Theorem~\ref{thm:RRlogloss} and Theorem~\ref{thm:main} we deduce relative loss bounds on the
logarithmic loss of kernelized Bayesian Ridge Regression and on the square
loss of kernelized Ridge Regression in comparison with the loss of any function
from a reproducing kernel Hilbert space. Both bounds have the form of equalities.

There is a lot of research done to prove upper and lower relative loss bounds under different loss functions. If the outcomes are assumed to be bounded, the strongest known theoretical guarantees for square loss are given by \citet{VovkCOS} and \citet{Azoury2001} for the algorithm which we call VAW (Vovk-Azoury-Warmuth) following~\citet{CesaBianchi2006}. 
In the case when the inputs and outcomes are not restricted in any way, like for our main guarantees, it is possible to prove certain loss bounds for the Gradient Descent;
see \citet{CesaBianchi1996}.

In Section~\ref{sec:protocol} of this paper we present the online regression framework and the main theoretical guarantee on the square loss of Ridge Regression. Section~\ref{sec:BA} describes what we call the Bayesian Algorithm.
In Section~\ref{sec:RRcompete} we show that Bayesian Ridge Regression is competitive with the experts which take into account the presence of noise in observations. In Section~\ref{sec:proofT} we prove the main theorem. Section~\ref{sec:kernelized} describes the case of infinite-dimensional Hilbert spaces.

\section{The prediction protocol and performance guarantees}\label{sec:protocol}
In online regression the learner follows this prediction protocol:
\begin{protocol}[H]
  \caption{Online regression protocol}
  \label{prot:OR}
  \begin{algorithmic}
    \FOR{$t=1,2,\dots$}
      \STATE Reality announces $x_t \in \R^n$
      \STATE Learner predicts $\gamma_t \in \R$
      \STATE Reality announces $y_t \in \R$
    \ENDFOR
  \end{algorithmic}
\end{protocol}

We use the Ridge Regression algorithm for the learner:
\begin{algorithm}[H]
    \caption{Online Ridge Regression}
    \label{alg:RR}
    \begin{algorithmic}
        \REQUIRE $a > 0$
        \STATE Initialize $b_0 = 0 \in \R^n, A_0 = aI \in \R^{n \times n}$
        \FOR{$t=1,2,\dots$}
            \STATE Read $x_t \in \R^n$
            \STATE Predict $\gamma_t = b_{t-1}'A_{t-1}^{-1}x_t$
            \STATE Read $y_t$
            \STATE Update $A_t = A_{t-1} + x_t x_t'$
            \STATE Update $b_t = b_{t-1} + y_t x_t$
        \ENDFOR
    \end{algorithmic}
\end{algorithm}
Following this algorithm the learner's prediction at step $T$ can be written as
\begin{equation*}
\gamma_T = \left(\sum_{t=1}^{T-1} y_t x_t \right)'\left(aI + \sum_{t=1}^{T-1} x_t x_t'\right)^{-1} x_T.
\end{equation*}
The incremental update of the matrix $A_t^{-1}$ can be done effectively by the Sherman-Morrison formula.
We prove the following theoretical guarantee for the square loss of the learner following Ridge Regression.
\begin{theorem}\label{thm:main}
  The Ridge Regression algorithm for the learner with $a > 0$ satisfies, at any step $T$,
  \begin{equation}\label{eq:main}
  \sum_{t=1}^T \frac { (y_t - \gamma_t)^2}{ 1+x_t'A_{t-1}^{-1}x_t} = \min_{\theta \in \R^n} \left(\sum_{t=1}^T (y_t - \theta' x_t)^2 + a\|\theta\|^2 \right).
  \end{equation}
\end{theorem}
Note that the part $x_t'A_{t-1}^{-1}x_t$ in the denominator is usually close to zero for large $t$.
An equivalent equality is also obtained (but well hidden) in the proof of Theorem 4.6 in \citet{Azoury2001}.
Our proof is more elegant.
We describe it from the point of view of online prediction, but we note the connection with Bayesian learning in derivations. We obtain an upper bound in the form which is more familiar from online prediction literature.
\begin{corollary}\label{cor:upperbound}
  Assume $|y_t| \le Y$ for all $t$, clip the predictions of Ridge Regression to $[-Y,Y]$, and denote them by $\gamma_t^Y$. Then
  \begin{equation}\label{eq:upperbound}
  \sum_{t=1}^T (y_t - \gamma_t^Y)^2 \le \min_\theta \left(\sum_{t=1}^T (y_t - \theta' x_t)^2 + a\|\theta\|^2 \right) + 4Y^2 \ln \det \left(I + \frac{1}{a} \sum_{t=1}^T x_t x_t'\right).
  \end{equation}
\end{corollary}
\begin{proof}
  We first clip the predictions of Ridge Regression to $[-Y,Y]$ in Theorem~\ref{thm:main}. In this case the loss at each step can only become smaller, and so the equality transforms to an inequality. Since all the outcomes also lie in $[-Y,Y]$, the maximum square loss at each step is $4Y^2$. We have the following relations:
  \begin{equation*}
  \frac {1}{ 1+x_t'A_{t-1}^{-1}x_t} = 1 - \left(\frac{x_t'A_{t-1}^{-1}x_t}{1 + x_t'A_{t-1}^{-1}x_t}\right) \text{ and } \frac{x_t'A_{t-1}^{-1}x_t}{1 + x_t'A_{t-1}^{-1}x_t} \le \ln(1+ x_t'A_{t-1}^{-1}x_t).
  \end{equation*}
  The last inequality holds because $x_t'A_{t-1}^{-1}x_t$ is non-negative due to the positive definiteness of the matrix $A_{t-1}$. Thus we can use $\frac{b}{1+b} \le \ln (1+b), b \ge 0$ (it holds at $b=0$, then take the derivatives of both sides). For the equality $\sum_{t=1}^T \ln (1+ x_t'A_{t-1}^{-1}x_t) = \ln \det \left(I + \frac{1}{a} \sum_{t=1}^T x_t x_t'\right)$ see~\eqref{eq:deteval}.
\end{proof}

The bound (\ref{eq:upperbound}) is exactly the bound obtained in Theorem~4 in \citet{VovkCOS} for the algorithm merging linear experts with predictions clipped to $[-Y,Y]$, which does not have a closed-form description and so is less interesting than clipped Ridge Regression. The bound for the VAW algorithm obtained in Theorem~1 in \citet{VovkCOS} has $Y^2$ in place of $4Y^2$ (the VAW algorithm is very similar to Ridge Regression; its predictions are $b'_{t-1}A_t^{-1}x_t$ rather than $b'_{t-1}A_{t-1}^{-1}x_t$). The regret term in (\ref{eq:upperbound}) has the logarithmic order in $T$ if $\|x_t\|_\infty \le X$ for all $t$, because
\begin{equation}\label{eq:determinant}
  \ln \det \left(I + \frac{1}{a} \sum_{t=1}^T x_t x_t'\right)
  \le n\ln \left(1 + \frac{TX^2}{a}\right)
\end{equation}
(the determinant of a positive definite matrix
is bounded by the product of its diagonal elements;
see Chapter~2, Theorem~7 of \citet{Beckenbach1961}.
This bound is also obtained in Theorem 4.6 in \citet{Azoury2001}.

From our Theorem~\ref{thm:main} we can also deduce Theorem~11.7 of \citet{CesaBianchi2006}, which is somewhat similar to our corollary. That theorem implies (\ref{eq:upperbound}) when Ridge Regression's predictions happen to be in $[-Y,Y]$ without clipping (but this is not what Corollary~\ref{cor:upperbound} asserts).

The upper bound~\eqref{eq:upperbound} does not hold if the coefficient $4$ is replaced by any number less than $\frac{3}{2\ln 2} \approx 2.164$, as can be seen from an example given in Theorem~3 in \citet{VovkCOS}, where the left-hand side of (\ref{eq:upperbound}) is $4T+o(T)$, the minimum in the right-hand side is at most $T$, $Y=1$, and the logarithm is $2T\ln 2+O(1)$. It is also known that there is no algorithm achieving~\eqref{eq:upperbound} with the coefficient less than $1$ instead of $4$ even in the case where $\left\|x_t\right\|_{\infty}\le X$ for all $t$; see Theorem~2 in \citet{VovkCOS}.

It is also possible to prove an upper bound without the logarithmic part on the cumulative square loss of Ridge Regression without assuming that the outcomes are bounded.
\begin{corollary}\label{cor:noregret}
  If $\|x_t\|_2 \le Z$ for all $t$ then
  the Ridge Regression algorithm for the learner with $a > 0$ satisfies, at any step $T$,
  \begin{equation}\label{eq:noregret}
    \sum_{t=1}^T (y_t - \gamma_t)^2 \le \left(1+\frac{Z^2}{a}\right) \min_{\theta \in \R^n} \left(\sum_{t=1}^T (y_t - \theta' x_t)^2 + a\|\theta\|^2 \right).
  \end{equation}
\end{corollary}
\begin{proof}
\citet{Qazaz1997} showed that $1 + x_t'A_{j}^{-1}x_t \le 1 + x_t'A_{i}^{-1}x_t$ for $j \ge i$. We take $i=0$ and obtain $1 + x_t'A_{t-1}^{-1}x_t \le 1 + Z^2/a$ for any $t$.
\end{proof}

This bound is better than the bound in Corollary~3.1 of \citet{Kakade2004}, which has an additional regret term of logarithmic order in time.

Asymptotic properties of the Ridge Regression algorithm
can be further studied using Corollary~A.1 in \citet{Kumon2009}.
It states that when $\|x_t\|_2 \le 1$ for all $t$,
then $x_t'A_{t-1}^{-1}x_t \to 0$ as $t \to \infty$.
It is clear that we can replace $\|x_t\|_2 \le 1$ for all $t$
by $\sup_t\|x_t\|_2 < \infty$.
The following corollary states that
if there exists a very good expert (asymptotically),
then Ridge Regression also predicts very well.
If there is no such a good expert,
Ridge Regression performs asymptotically as well as the best regularized expert.
\begin{corollary}\label{cor:asympotic}
  Let $a>0$ and $\gamma_t$ be the predictions output by the Ridge Regression algorithm
  with parameter $a$.
  Suppose $\sup_t\left\|x_t\right\|_2<\infty$.
  \begin{enumerate}
  \item
    If
    \begin{equation}\label{eq:condition-1}
      \exists\theta\in\R^n:
      \sum_{t=1}^{\infty}
      (y_t - \theta' x_t)^2
      <
      \infty,
    \end{equation}
    then
    $$
      \sum_{t=1}^{\infty}
      (y_t - \gamma_t)^2
      <
      \infty.
    $$
  \item
    If
    \begin{equation}\label{eq:condition-2}
      \forall\theta\in\R^n:
      \sum_{t=1}^{\infty}
      (y_t - \theta' x_t)^2
      =
      \infty,
    \end{equation}
    then
    \begin{equation}\label{eq:conclusion}
      \lim_{T\to\infty}
      \frac
      {
        \sum_{t=1}^T
        (y_t - \gamma_t)^2
      }
      {
        \min_{\theta\in\R^n}
        \left(
          \sum_{t=1}^T
	  (y_t - \theta' x_t)^2 + a\|\theta\|^2
        \right)
      }
      =
      1.
    \end{equation}
  \end{enumerate}
\end{corollary}
\begin{proof}
\textbf{Part~1.}
  Suppose that the condition~\eqref{eq:condition-1} holds.
  Then the right-hand side of~\eqref{eq:main}
  is bounded by a constant (independent of $T$).
  By Corollary~A.1 in \citet{Kumon2009},
  the denominators in the left-hand side converge to $1$ as $t\to\infty$
  and so are bounded.
  Therefore,
  the sequence $\sum_{t=1}^T(y_t-\gamma_t)^2$ remains bounded as $T\to\infty$.

  \textbf{Part~2.}
  Suppose that the condition~\eqref{eq:condition-2} holds
  and the right-hand side of~\eqref{eq:main}
  is bounded above by a constant $C$.
  Then for each $T$ there exists $\theta_T$ such that
  $$
    \sum_{t=1}^T
    (y_t - \theta'_T x_t)^2
    +
    a\left\|\theta_T\right\|^2
    \le
    C.
  $$
  It follows that each $\theta_T$ belongs
  to the closed ball with centre $0$ and of radius $\sqrt{C/a}$.
  This ball is a compact set, and thus
  the sequence $\theta_T$ has a subsequence that converges to some $\tilde \theta$.
  For each $T_0$
  we have $\sum_{t=1}^{T_0} (y_t-\tilde \theta' x_t)^2 \le C$,
  because otherwise we would have $\sum_{t=1}^{\hat T} (y_t - \theta'_{\hat T} x_t)^2 > C$
  for a large enough $\hat T$ in the subsequence.
  Therefore, we have arrived at a contradiction:
  $\sum_{t=1}^{\infty} (y_t-\tilde \theta' x_t)^2 \le C<\infty$.

  Once we know that the right-hand side of~\eqref{eq:main}
  tends to $\infty$ as $T\to\infty$
  and the denominators on the left-hand side tend to $1$
  (this is true by Corollary~A.1 in \citealp{Kumon2009}),
  (\ref{eq:conclusion}) becomes intuitively plausible
  since, as far as the conclusion (\ref{eq:conclusion}) is concerned,
  we can ignore the finite number of $t$s
  for which the denominator $1+x'_tA_{t-1}^{-1}x_t$ is significantly different from $1$.
  We will, however, give a formal argument.

  The inequality ${}\ge1$ in (\ref{eq:conclusion})
  is clear from~\eqref{eq:main} and $1+x'_tA_{t-1}^{-1}x_t \ge 1$.
  We shall prove the inequality ${}\le1$ now.
  Choose a small $\epsilon>0$.
  Then starting from some $t = T_0$
  we have that the denominators $1+x'_tA_{t-1}^{-1}x_t$
  are less than $1+\epsilon$.
  Thus, for $T>T_0$,
  \begin{multline*}
    \sum_{t=1}^T
    (y_t-\gamma_t)^2
    =
    \sum_{t=1}^{T_0}
    (y_t-\gamma_t)^2
    +
    \sum_{t=T_0+1}^T
    (y_t-\gamma_t)^2
    \\ \le
    \sum_{t=1}^{T_0}
    (y_t-\gamma_t)^2
    +
    (1+\epsilon)
    \sum_{t=1}^T
    \frac{(y_t-\gamma_t)^2}{1+x'_tA_{t-1}^{-1}x_t}\\
    =
    \sum_{t=1}^{T_0}
    (y_t-\gamma_t)^2
    +
    (1+\epsilon)
    \min_{\theta\in\R^n}
    \left(
      \sum_{t=1}^T
      (y_t-\theta' x_t)^2
      +
      a\left\|\theta\right\|^2
    \right).
  \end{multline*}
  This implies that the left-hand side of (\ref{eq:conclusion})
  with $\lim$ replaced by $\limsup$
  does not exceed $1+\epsilon$,
  and it remains to remember that $\epsilon$
  can be taken arbitrarily small.
\end{proof}

\section{Bayesian algorithm}\label{sec:BA}
In this section we describe the main algorithm used to prove our theoretical bounds. Let us denote the set of possible outcomes by $\Omega$, the index set for the experts by $\Theta$, and the set of allowed predictions by $\Gamma$. The quality of predictions is measured by a loss function $\lambda: \Gamma \times \Omega \to \R$. We have $\Omega = \R$, $\Theta = \R^n$, and $\Gamma$ is the set of all measurable functions on the real line integrable to one. The loss function $\lambda$ is the logarithmic loss $\lambda(\gamma,y) = -\ln \gamma(y)$, where $\gamma \in \Gamma$ and $y \in \Omega$.
The learner follows the prediction with expert advice protocol.
\begin{protocol}[H]
  \caption{Prediction with expert advice protocol}
  \label{prot:PEA}
  \begin{algorithmic}
    \STATE Initialize $L_0:=0$ and $L_0(\theta) = 0$, $\forall \theta\in\Theta$
    \FOR{$t=1,2,\dots$}
      \STATE Experts $\theta\in\Theta$ announce their predictions $\xi_t^\theta \in \Gamma$
      \STATE Learner predicts $\gamma_t \in \Gamma$
      \STATE Reality announces $y_t \in \Omega$
      \STATE Losses are updated: $L_T = L_{T-1} + \lambda(\gamma_t,y_t)$, $L_T(\theta) = L_{T-1}(\theta) + \lambda(\xi_t^\theta,y_t), \forall \theta \in \Theta$
    \ENDFOR
  \end{algorithmic}
\end{protocol}
\noindent
Here by $L_T$ we denote the cumulative loss of the learner at step $T$, and by $L_T(\theta)$ we denote the cumulative loss of the expert $\theta$ at this step.

We use a standard algorithm in prediction with expert advice (a special case of the Aggregating Algorithm for the logarithmic loss function and learning rate 1, going back to \citet{DeSantis1988} in the case of countable $\Theta$ and $\Omega$) to derive the main theoretical bound and give predictions. We call it the Bayesian Algorithm (BA) as it is virtually identical to the Bayes rule used in Bayesian learning (the main difference being that the experts are not required to follow any prediction strategies). Instead of looking for the best expert, the algorithm considers all the experts and takes a weighted average of their predictions as its own prediction. In detail, it works as follows.
\begin{algorithm}[H]
    \caption{Bayesian Algorithm}
    \label{alg:BA}
    \begin{algorithmic}
        \REQUIRE A probability measure $P_0(d\theta) = P_0^*(d\theta)$ on $\Theta$ (the prior distribution, or weights)
        \FOR{$t=1,2,\dots$}
            \STATE Read experts' predictions $\xi_t^\theta \in \Gamma, \forall \theta \in \Theta$
            \STATE Predict $g_t = \int_\Theta \xi_t^\theta P_{t-1}^*(d\theta)$
            \STATE Read $y_t$
            \STATE Update the weights $P_t(d\theta) = \xi_t^\theta(y_t) P_{t-1}(d\theta)$
            \STATE Normalize the weights $P_t^*(d\theta) = P_t(d\theta)/\int_\Theta P_t(d\theta)$
        \ENDFOR
    \end{algorithmic}
\end{algorithm}

The experts' weights are updated according to their losses at each step:
$
\xi_t^\theta(y_t) = e^{-\lambda(\xi_t^\theta,y_t)}
$;
larger losses lead to smaller weights.
After $t$ steps the weights become
\begin{equation}\label{eq:weights}
  P_t(d\theta) = e^{-L_t(\theta)} P_0(d\theta).
\end{equation}
The normalized weights $P_T^*(d\theta)$ correspond to the posterior distribution over $\theta$ after the step $T$.
As we said,
the prediction of the BA at step $T$ is given by the average
\begin{equation}\label{eq:BAprediction}
g_T = \int_\Theta \xi_T^\theta P_{T-1}^*(d\theta)
\end{equation}
of the experts' predictions.

The next lemma is a special case of Lemma~1 in \citet{VovkCOS}.
It shows that the cumulative loss of the BA is an average of the experts' cumulative losses in a generalized sense, as in, e.g., Chapter 3 of \citet{Hardy1952}.
\begin{lemma}\label{lem:lossAPA}
  For any prior $P_0$ and any $T=1,2,\ldots$, the cumulative loss of the BA can be expressed as
  \begin{equation}\label{eq:lossAPA}
    L_T
    =
    -\ln \int_{\Theta} e^{- L_T(\theta)} P_0(d\theta).
  \end{equation}
\end{lemma}
\begin{proof}
  We proceed by induction in $T$:
  for $T=0$ the equality is obvious,
  and for $T>0$ we have:
  \begin{multline*}
    L_T
    =
    L_{T-1}
    -
    \ln g_T(y_T)
    \\ =
    -\ln \int_{\Theta} e^{- L_{T-1}(\theta)} P_0(d\theta)
    -
    \ln
    \int_{\Theta}
      \xi^{\theta}_T
      \frac{e^{- L_{T-1}(\theta)}}{\int_{\Theta} e^{- L_{T-1}(\theta)} P_0(d\theta)}
    P_0(d\theta)\\
    =
    -\ln \int_{\Theta} e^{- L_T(\theta)} P_0(d\theta)
  \end{multline*}
  (the second equality follows from the inductive assumption, the definition of $g_T$,
  and (\ref{eq:weights})).
\end{proof}

\section{Bayesian Ridge Regression as a competitive algorithm}\label{sec:RRcompete}
Let us consider experts whose predictions at step $t$ are the densities of the normal distributions $N(\theta' x_t, \sigma^2)$ on the set of outcomes for some fixed variance $\sigma^2>0$ (so each expert $\theta$ follows a fixed strategy). From the statistical point of view, they predict according to the model $y_t = \theta' x_t + \epsilon_t$ with Gaussian noise $\epsilon_t\sim N(0,\sigma^2)$. In other words, the prediction of each expert $\theta \in \Theta$ is
\begin{equation}\label{eq:expert}
\xi_t^\theta(y) = \frac{1}{\sqrt{2\pi\sigma^2}}e^{-\frac{(y - \theta' x_t)^2}{2\sigma^2}}.
\end{equation}

Let us take the initial distribution $N(0,\frac{\sigma^2}{a}I)$ on the experts with some ${a > 0}$:
\begin{equation*}
P_0(d\theta) = \left(\frac{a}{2\sigma^2\pi}\right)^{n/2}\exp\left(-\frac{a}{2\sigma^2}\|\theta\|^2\right)d\theta.
\end{equation*}
We will prove that in this setting the prediction of the Bayesian Algorithm is equal to the prediction of Bayesian Ridge Regression. But first we need to introduce some notation. For $t\in\{1,2,\ldots\}$, let $X_t$ be the $t\times n$ matrix of row vectors $x_1',\ldots,x_t'$ and $Y_t$ be the column vector of outcomes $y_1,\ldots,y_t$. Let $A_t = X'_tX_t + aI$, as before. Bayesian Ridge Regression is the algorithm predicting at each step $T$ the normal distribution $N(\gamma_T,\sigma_T^2)$ with the mean and variance given by
\begin{equation}\label{eq:RRBayes}
\gamma_T = Y'_{T-1} X_{T-1} A_{T-1}^{-1} x_T , \quad \sigma_T^2 = \sigma^2 x_T' A_{T-1}^{-1} x_T + \sigma^2
\end{equation}
for some $a > 0$ and the known noise variance $\sigma^2$.
\begin{lemma}\label{lem:RRBayesian}
  In our setting the prediction \eqref{eq:BAprediction} of the Bayesian Algorithm is the prediction density of Bayesian Ridge Regression in the notation of~\eqref{eq:RRBayes}:
  \begin{equation}\label{eq:RRBayesian}
  g_T(y) = \frac{1}{\sqrt{2\pi\sigma_T^2}} e^{-\frac{(y-\gamma_T)^2}{2\sigma_T^2}}.
  \end{equation}
\end{lemma}
\begin{proof}
  The prediction
  $$
    g_T(y)
    =
    \int_\Theta \xi_T^\theta(y) P_{T-1}^*(d\theta)
    =
    \frac
    {
      \int_{\R^n}
      \frac{1}{\sqrt{2\pi\sigma^2}} e^{-\frac{(y-\theta'x_T)^2}{2\sigma^2}}
      \prod_{t=1}^{T-1}
      \frac{1}{\sqrt{2\pi\sigma^2}} e^{-\frac{(y_t-\theta'x_t)^2}{2\sigma^2}}
      P_0(d\theta)
    }
    {
      \int_{\R^n}
      \prod_{t=1}^{T-1}
      \frac{1}{\sqrt{2\pi\sigma^2}} e^{-\frac{(y_t-\theta'x_t)^2}{2\sigma^2}}
      P_0(d\theta)
    }
  $$
  formally coincides with the density of the predictive distribution of the Bayesian Gaussian linear model, and so equality~\eqref{eq:RRBayesian} is true: see Section~3.3.2 of \citet{Bishop2006}.
\end{proof}

\begin{remark}
  From the probabilistic point of view Lemma~\ref{lem:RRBayesian} is usually explained in the following way \citep{Hoerl2000}. The posterior distribution $P_{T-1}^*(\theta)$ is $N(A_{T-1}^{-1} X_{T-1}'Y_{T-1},\sigma^2 A_{T-1}^{-1})$. The conditional distribution of $\theta' x_T$ given the training examples is then $N(Y'_{T-1} X_{T-1} A_{T-1}^{-1} x_T, \sigma^2 x_T'A_{T-1}^{-1}x_T)$, and so the predictive distribution is $N(Y'_{T-1} X_{T-1} A^{-1}_{T-1} x_T, \sigma^2 x_T' A_{T-1}^{-1} x_T + \sigma^2)$.
\end{remark}

For the subsequent derivations, we will need the following well-known lemma, whose proof can be found in Lemma~ 8 of \citet{Busuttil2008} or extracted from Chapter~2, Theorem~3~of \citet{Beckenbach1961}.
\begin{lemma}\label{lem:inteval}
  Let $W(\theta) = \theta' A \theta + b'\theta +c$ for $\theta,b \in \R^n$, $c$ be a scalar, and $A$ be a symmetric positive definite $n \times n$ matrix. Then
  \begin{equation*}
  \int_{\R^n} e^{-W(\theta)} d\theta = e^{-W_0} \frac{\pi^{n/2}}{\sqrt{\det A}},
  \end{equation*}
  where $W_0 = \min_\theta W(\theta)$.
\end{lemma}

The right-hand side of (\ref{eq:lossAPA}) can be transformed to the regularized cumulative loss of the best expert $\theta$ and a regret term:
\begin{theorem}\label{thm:RRlogloss}
  For any sequence $x_1,y_1,x_2,y_2,\ldots,$ the cumulative logarithmic loss of the Bayesian Ridge Regression algorithm~\eqref{eq:RRBayesian} at any step $T$ can be expressed as
  \begin{equation}\label{eq:RRlogloss}
  L_T = \min_\theta \left(L_T(\theta) + \frac{a}{2\sigma^2}\|\theta\|^2\right) + \frac{1}{2}\ln \det \left( I + \frac{1}{a}\sum_{t=1}^T x_t x_t' \right).
  \end{equation}
  If $\|x_t\|_\infty \le X$ for any $t = 1,2,\ldots,$ then
  \begin{equation}\label{eq:RRlogloss-simplified}
  L_T \le \min_\theta \left(L_T(\theta) + \frac{a}{2\sigma^2}\|\theta\|^2\right) + \frac{n}{2}\ln \left( 1 + \frac{TX^2}{a}\right).
  \end{equation}
\end{theorem}
\begin{proof}
  We have to calculate the right-hand side of~(\ref{eq:lossAPA}). The integral is expressed as
  \begin{equation*}
  \int_\Theta \frac{1}{(2\pi\sigma^2)^{T/2}}\left(\frac{a}{2\sigma^2\pi}\right)^{n/2}e^{-\frac{1}{2\sigma^2}\left(\sum_{t=1}^T(y_t - \theta' x_t)^2 + a\|\theta\|^2 \right)}d\theta.
  \end{equation*}
  By Lemma~\ref{lem:inteval} it is equal to
  \begin{equation*}
  \frac{1}{(2\pi\sigma^2)^{T/2}} \left(\frac{a}{2\sigma^2\pi}\right)^{n/2} e^{-\frac{1}{2\sigma^2} \left(\sum_{t=1}^T (y_t - \theta'_0 x_t)^2 + a \|\theta_0\|^2 \right) } \frac{\pi^{n/2}}{\sqrt{\det A_T}},
  \end{equation*}
  where $A_T$ is the coefficient matrix in the quadratic part: $A_T = \frac{1}{2\sigma^2}(aI + \sum_{t=1}^T x_t x_t')$ and $\theta_0$ is the best predictor: $\theta_0 = \arg\min_{\theta}\left(\sum_{t=1}^T (y_t - \theta' x_t)^2 +a \|\theta\|^2\right)$. Taking the minus logarithm of this expression we get
  \begin{equation*}
  -\sum_{t=1}^T \ln\left(\frac{1}{\sqrt{2\pi\sigma^2}} e^{-\frac{1}{2\sigma^2}(y_t - \theta'_0 x_t)^2}\right) + \frac{a}{2\sigma^2}\|\theta_0\|^2 + \frac{1}{2}\ln \det \left(I + \frac{1}{a}\sum_{t=1}^T x_t x_t'\right).
  \end{equation*}
  To obtain the upper bound (\ref{eq:RRlogloss-simplified})
  it suffices to apply (\ref{eq:determinant}).
\end{proof}

This theorem shows that the Bayesian Ridge Regression algorithm can be thought of as an online algorithm successfully competing with all the Gaussian linear models under the logarithmic loss function. Similar bounds on the logarithmic loss of Bayesian Ridge Regression are proven by~\citet{Kakade2004}.

\section{Proof of Theorem~\ref{thm:main}}\label{sec:proofT}
  Let us rewrite $L_T$ and $L_T(\theta)$ using \eqref{eq:RRBayesian},
  the expression for $\sigma_t^2$ given by~\eqref{eq:RRBayes},
  and (\ref{eq:expert}):
  \begin{align*}
  L_T
  &=
  -\sum_{t=1}^T
  \ln \left(\frac{1}{\sqrt{2\pi\sigma_t^2}} e^{-\frac{(y-\gamma_t)^2}{2\sigma_t^2}}\right)
  \\ &=
\frac{1}{2}\ln \left((2\pi\sigma^2)^T \prod_{t=1}^T (1 + x_t'A_{t-1}^{-1} x_t)\right) + \frac{1}{2\sigma^2 }\sum_{t=1}^T \frac{(y_t - \gamma_t)^2}{1 + x_t'A_{t-1}^{-1} x_t},\\
  L_T(\theta)
  &=
  \sum_{t=1}^T \lambda(\xi_t^\theta,y_t)
  =
  -\ln \left(\frac{1}{(2\pi\sigma^2)^{T/2}}e^{-\frac{1}{2\sigma^2}\sum_{t=1}^T(y_t - \theta' x_t)^2}\right)
  \\ &= \frac{T}{2}\ln (2\pi\sigma^2) + \frac{1}{2\sigma^2}\sum_{t=1}^T (y_t - \theta' x_t)^2.
  \end{align*}
  Substituting these expression into~\eqref{eq:RRlogloss} we have:
  \begin{multline*} \frac{1}{2}\ln \prod_{t=1}^T (1 + x_t'A_{t-1}^{-1} x_t) + \frac{1}{2\sigma^2 }\sum_{t=1}^T \frac{(y_t - \gamma_t)^2}{1 + x_t'A_{t-1}^{-1} x_t}\\
  = \frac{1}{2\sigma^2}
  \min_\theta\left(\sum_{t=1}^T (y_t - \theta' x_t)^2 + a\|\theta\|^2\right) + \frac{1}{2}\ln \det \left( I + \frac{1}{a}\sum_{t=1}^T x_t x_t' \right).
  \end{multline*}
  Equation~(\ref{eq:main}) follows from the fact that
  \begin{equation}\label{eq:deteval}
  \det \left( I + \frac{1}{a}\sum_{t=1}^T x_t x_t' \right)
  = \prod_{t=1}^T (1 + x_t'A_{t-1}^{-1} x_t)
  \end{equation}
  for $A_t = aI + \sum_{i=1}^t x_i x_i'$. This fact can be proven by induction in $T$: for $T=0$ it is obvious ($1=1$) and for $T \ge 1$ we have
  \begin{multline*}
  \det \left( I + \frac{1}{a}\sum_{t=1}^T x_t x_t' \right)
  =
  a^{-n}
  \det A_T = a^{-n} \det \left(A_{T-1}  + x_T x_T' \right)
  \\ =
  a^{-n}
  (1+x_T'A_{T-1}^{-1}x_T)\det A_{T-1}
  =
  \det \left( I + \frac{1}{a}\sum_{t=1}^{T-1} x_t x_t' \right)
  (1+x_T'A_{T-1}^{-1}x_T)
  \\ =
  \prod_{t=1}^T
  (1+x_t'A_{t-1}^{-1}x_t).
  \end{multline*}
  The third equality follows from the Matrix Determinant Lemma: see, e.g., Theorem~18.1.1~of \citet{Harville1997}. The last equality follows from the inductive assumption.
  Note that $\sigma^2$ canceled out;
  this is natural as Ridge Regression (unlike Bayesian Ridge Regression)
  does not depend on $\sigma$.

\section{Kernelized Ridge Regression}\label{sec:kernelized}
In this section we prove bounds on the square loss of kernelized Ridge Regression. We also prove bounds on the logarithmic loss for a commonly used non-parametric Gaussian algorithm: kernelized Bayesian Ridge Regression.
These bounds explicitly handle infinite dimensional classes of experts.

Let $\mathbf{X}$ be an arbitrary set of inputs. We define a \emph{reproducing kernel Hilbert space (RKHS)} $\mathcal{F}$ of functions $\mathbf{X} \to \R$ as a functional Hilbert space with continuous evaluation functional $f \in \mathcal{F} \mapsto f(x)$ for each $x \in \mathbf{X}$. By the Riesz-Fischer theorem for any $x \in \mathbf{X}$ there is a unique $k_x \in \mathcal{F}$ such that $\langle k_x, f \rangle_\mathcal{F} = f(x)$ for any $f \in \mathcal{F}$. The \emph{kernel} $\mathcal{K}: \mathbf{X}^2 \to \R$ of the RKHS $\mathcal{F}$ is defined as $\mathcal{K}(x_1,x_2) = \langle k_{x_1}, k_{x_2} \rangle$ for any $x_1,x_2 \in \mathbf{X}$. For more information about kernels please refer to \citet{Scholkopf2002}.

Let us introduce some notation. Let $\mathbf{K}_t$ be the kernel matrix $\mathcal{K}(x_i,x_j)$ at step $t$, where ${i,j = 1,\ldots,t}$. Let $\mathbf{k}_t$ be the column vector $\mathcal{K}(x_i,x_t)$ for $i=1,\ldots,t-1$. As before, $Y_t$ is the column vector of outcomes $y_1,\ldots,y_t$. The kernelized Ridge Regression is defined as the learner's strategy in Protocol \ref{prot:OR} that predicts $\gamma_T = Y'_{T-1}(aI + \mathbf{K}_{T-1})^{-1}\mathbf{k}_T$ at each step $T$; see, e.g., \citet{Saunders1998}.
The following theorem 
is an analogue of Theorem~\ref{thm:main} for kernelized Ridge Regression;
in its proof we will see how kernelized Ridge Regression is connected with Ridge Regression.
\begin{theorem}\label{thm:mainkernel}
  The kernelized Ridge Regression algorithm for the learner with $a > 0$ satisfies, at any step $T$,
  \begin{equation}\label{eq:mainkernel}
  \sum_{t=1}^T \frac { (y_t - \gamma_t)^2}{ 1+(\mathcal{K}(x_t,x_t) - \mathbf{k}'_t(aI + \mathbf{K}_{t-1})^{-1}\mathbf{k}_t)/a} = \min_{f \in \mathcal{F}} \left(\sum_{t=1}^T (y_t - f(x_t))^2 + a\|f\|^2_\mathcal{F} \right).
  \end{equation}
\end{theorem}
\begin{proof}
  It suffices to prove that for each $T\in\{1,2,\ldots\}$
  and every sequence of input vectors and outcomes
  $(x_1,y_1,\ldots,x_T,y_T)\in(\mathbf{X}\times\R)^T$
  the equality (\ref{eq:mainkernel}) is satisfied.
  Fix such $T$ and $(x_1,y_1,\ldots,x_T,y_T)$;
  our goal is to prove (\ref{eq:mainkernel}).
  Fix an isomorphism between the linear span of $k_{x_1},\ldots,k_{x_T}$
  and $\R^{\tilde T}$,
  where $\tilde T\le T$ is the dimension of the linear span of $k_{x_1},\ldots,k_{x_T}$.
  Let $\tilde x_1,\ldots,\tilde x_T\in\R^{\tilde T}$ be the images of $k_{x_1},\ldots,k_{x_T}$,
  respectively,
  under this isomorphism.
  Notice that, for all $t$,
  $\mathbf{K}_t$ is the matrix $\langle \tilde x_i, \tilde x_j\rangle$,
  $i,j = 1,\ldots,t$,
  and $\mathbf{k}_t$ is the column vector $\langle \tilde x_i,\tilde x_t\rangle$ for $i=1,\ldots,t-1$.
  We know that (\ref{eq:main}) with $\tilde x_t$ in place of $x_t$
  and $\tilde\gamma_t$ in place of $\gamma_t$
  holds for Ridge Regression,
  whose predictions are now denoted $\tilde\gamma_t$
  (in order not to confuse them with kernelized Ridge Regression's predictions $\gamma_t$).
  The predictions output by Ridge Regression on $\tilde x_1,y_1,\ldots,\tilde x_T,y_T$
  and by kernelized Ridge Regression on $x_1,y_1,\ldots,x_T,y_T$
  are the same:
  \begin{multline*}
    \gamma_t
    =
    Y'_{t-1}(aI + \mathbf{K}_{t-1})^{-1}\mathbf{k}_t
    =
    Y'_{t-1}(aI + \tilde X_{t-1}\tilde X'_{t-1})^{-1}\tilde X_{t-1} \tilde x_t
    \\ =
    Y'_{t-1}\tilde X_{t-1}(aI + \tilde X'_{t-1} \tilde X_{t-1})^{-1} \tilde x_t
    =
    \tilde\gamma_t
  \end{multline*}
  (for the notation see~\eqref{eq:RRBayes}, with tildes added).
  The denominators in (\ref{eq:mainkernel}) and (\ref{eq:main}) are also the same:
  \begin{multline*}
    1 + (\mathcal{K}(x_t,x_t) - \mathbf{k}'_t (aI + \mathbf{K}_{t-1})^{-1} \mathbf{k}_t) / a
    \\ =
    1 + \tilde x'_t (I - \tilde X'_{t-1}(aI + \tilde X_{t-1} \tilde X'_{t-1})^{-1}
    \tilde X_{t-1}) \tilde x_t / a
    \\ =
    1 + \tilde x'_t (aI + \tilde X'_{t-1} \tilde X_{t-1})^{-1}
    ((aI + \tilde X'_{t-1} \tilde X_{t-1}) - \tilde X'_{t-1} \tilde X_{t-1}) \tilde x_t /a
    \\ =
    1 + \tilde x'_t (aI + \tilde X_{t-1}' \tilde X_{t-1})^{-1} \tilde x_t.
  \end{multline*}
  The right-hand sides are the same by the representer theorem
  (see, e.g., Theorem~4.2 in \citealp{Scholkopf2002}).
  Indeed, by this theorem we have
  \begin{multline*}
  \min_{f \in \mathcal{F}} \left(\sum_{t=1}^T (y_t - f(x_t))^2 + a\|f\|^2_\mathcal{F} \right) 
  \\ =
  \min_{c_1,\ldots,c_T \in \R} \left(\sum_{t=1}^T \left(y_t - \sum_{i=1}^T c_i \mathcal{K}(x_i,x_t)\right)^2 + a\left\|\sum_{i=1}^T c_i k_{x_i}\right\|^2_\mathcal{F} \right) \\
  = \min_{c_1,\ldots,c_T \in \R} \left(\sum_{t=1}^T \left(y_t - \sum_{i=1}^T c_i \langle \tilde x_i, \tilde x_t\rangle \right)^2 + a\left\|\sum_{i=1}^T c_i \tilde x_i \right\|^2_2 \right) \enspace
  \end{multline*}
  (the last equality holds due to the isomorphism).
  Denoting $\theta = \sum_{i=1}^T c_i \tilde x_i \in \R^{\tilde T}$
  we obtain the expression for the minimum in~\eqref{eq:main}:
  $\theta$ ranges over the whole of $\R^{\tilde T}$
  (as $c_1,\ldots,c_T$ range over $\R$)
  since $\tilde x_1,\ldots,\tilde x_T$ span $\R^{\tilde T}$.
\end{proof}

Similarly to the proof of Theorem~\ref{thm:mainkernel} we can prove an analogue of Theorem~\ref{thm:RRlogloss} for kernelized Bayesian Ridge Regression. At step $T$ kernelized Bayesian Ridge Regression predicts the normal density on outcomes with the mean $\gamma_T$ and variance $\sigma^2 + \sigma^2(\mathcal{K}(x_T,x_T) - \mathbf{k}'_T(aI + \mathbf{K}_{T-1})^{-1}\mathbf{k}_T)/a$. We denote by $L_T$ the cumulative logarithmic loss, over the first $T$ steps, of the algorithm and by $L_T(f)$ the cumulative logarithmic loss of the expert $f$ predicting normal density with the mean $f(x_t)$ and variance $\sigma^2$.
\begin{theorem}\label{thm:RRloglosskernel}
  For any sequence $x_1,y_1,x_2,y_2,\ldots,$ the cumulative logarithmic loss of the kernelized Bayesian Ridge Regression algorithm at any step $T$ can be expressed as
  \begin{equation*}
  L_T = \min_{f \in \mathcal{F}} \left(L_T(f) + \frac{a}{2\sigma^2}\|f\|^2_\mathcal{F}\right) + \frac{1}{2}\ln \det \left( I + \frac{1}{a}\mathbf{K}_T \right).
  \end{equation*}
\end{theorem}
\noindent
This theorem is proven by \citet{Kakade2005} for $a=1$.

We can see from Theorem~13.3.8 of \citet{Harville1997} that
\begin{multline*}
\det \left(I + \frac{1}{a}\mathbf{K}_T\right) =
\det \begin{pmatrix}
    I + \mathbf{K}_{T-1}/a & \mathbf{k}_{T}/a\\
    \mathbf{k}_{T}'/a & 1 + \mathcal{K}(x_T,x_T)/a
\end{pmatrix}\\
= \det \left(I + \frac{1}{a}\mathbf{K}_{T-1}\right) (1 + (\mathcal{K}(x_T,x_T) - \mathbf{k}_{T}'(aI + \mathbf{K}_{T-1})^{-1}\mathbf{k}_{T})/a),
\end{multline*}
and so by induction we have
\begin{equation*}
\det \left(I + \frac{1}{a}\mathbf{K}_T\right)
= \prod_{t=1}^T (1 + (\mathcal{K}(x_t,x_t) - \mathbf{k}_{t}'(aI + \mathbf{K}_{t-1})^{-1}\mathbf{k}_{t})/a),
\end{equation*}
with $\mathbf{k}'_1(aI+\mathbf{K}_0)^{-1}\mathbf{k}_1$ understood to be $0$.
Using this equality and following the arguments of the proof of Corollary~\ref{cor:upperbound} we obtain the following corollary from Theorem~\ref{thm:mainkernel}.
\begin{corollary}\label{cor:upperboundkernel}
  Assume $|y_t| \le Y$ for all $t$, clip the predictions of kernelized Ridge Regression to $[-Y,Y]$, and denote them by $\gamma_t^Y$. Then
  \begin{equation}\label{eq:upperboundkernel}
  \sum_{t=1}^T (y_t - \gamma_t^Y)^2 \le \min_{f \in \mathcal{F}} \left(\sum_{t=1}^T (y_t - f(x_t))^2 + a\|f\|^2_\mathcal{F} \right) + 4Y^2 \ln \det \left(I + \frac{1}{a} \mathbf{K}_T\right).
  \end{equation}
\end{corollary}
\noindent
It is possible to prove this corollary directly from Corollary~\ref{cor:upperbound} using the same argument as in the proof of Theorem~\ref{thm:mainkernel}.

The order of the regret term in~\eqref{eq:upperboundkernel} is not clear on the face of it. We show that it has the order $O(\sqrt{T})$ in many cases. We will use the notation $c^2_\mathcal{F} = \sup_{x \in \mathbf{X}} \mathcal{K}(x,x)$.
We bounding the logarithm of the determinant and obtain that
$
\ln\det \left(I + \frac{1}{a} \mathbf{K}_T\right) \le T \ln \left(1+\frac{c^2_{\mathcal{F}}}{a}\right)
$
(cf.\ (\ref{eq:determinant})).
If we know the number $T$ of steps in advance, then we can choose a specific value for $a$; let $a=c_{\mathcal{F}}\sqrt{T}$. Thus we get an upper bound with the regret term of the order $O(\sqrt{T})$ for any $f \in \mathcal{F}$:
\begin{equation*}
  \sum_{t=1}^T (y_t - \gamma_t^Y)^2 \le \sum_{t=1}^T (y_t - f(x_t))^2 + c_{\mathcal{F}}(\|f\|^2_\mathcal{F} + 4Y^2)\sqrt{T}.
\end{equation*}
If we do not know the number of steps in advance, it is possible to achieve a similar bound using the Bayesian Algorithm with a suitable prior over the parameter $a$:
\begin{multline}\label{eq:bounded-1}
  \sum_{t=1}^T (y_t - \gamma_t^Y)^2 \le \sum_{t=1}^T (y_t - f(x_t))^2 + 8Y\max\left(c_\mathcal{F} \|f\|_\mathcal{F},Y\delta T^{-1/2 + \delta}\right)\sqrt{T+2}\\
   + 6Y^2 \ln T + c_\mathcal{F}^2 \|f\|_\mathcal{F}^2 + O(Y^2)
\end{multline}
for any arbitrarily small $\delta > 0$, where the constant implicit in $O(Y^2)$ depends only on $\delta$. (Proof omitted.)

In particular,
\eqref{eq:bounded-1} shows that if $\mathbf{X}$ is a universal kernel \citep{Steinwart2001}
on a topological space $\mathbf{X}$,
Ridge Regression is competitive with all continuous functions on $\mathbf{X}$:
for any continuous $f:\mathbf{X}\to\R$,
\begin{equation}\label{eq:bounded-2}
  \limsup_{T\to\infty}
  \frac1T
  \left(
    \sum_{t=1}^T (y_t - \gamma_t^Y)^2
    -
    \sum_{t=1}^T (y_t - f(x_t))^2
  \right)
  \le
  0
\end{equation}
(assuming $\lvert y_t\rvert\le Y$ for all $t$).
For example, (\ref{eq:bounded-2}) holds for $\mathbf{X}$ a compact set in $\R^n$,
$\mathcal{K}$ an RBF kernel, and $f:\mathbf{X}\to\R$ any continuous function,
see Example~1 of \citet{Steinwart2001}.

\subsection*{Acknowledgements}
We are very grateful for useful comments to Yuri Kalnishkan and Alexey Chernov. Thanks to the organizers and lecturers of the Cambridge Machine Learning Summer School 2009, whose work helped us to look at the usual problems from a new viewpoint. This work was supported by EPSRC (grant EP/F002998/1).

\end{document}